# A Neural Network Decision Tree for Learning Concepts from EEG Data

Vitaly Schetinin
TheorieLabor, Friedrich-Schiller-Universität Jena, Germany
Vitaly.Schetinin@uni-jena.de, http://nnlab.tripod.com

***Abstract*** ***– To learn the multi-class conceptions from the electroencephalogram (EEG) data we developed a neural network decision tree (DT), that performs the linear tests, and a new training algorithm. We found that the known methods fail inducting the classification models when the data are presented by the features some of them are irrelevant, and the classes are heavily overlapped. To train the DT, our algorithm exploits a bottom up search of the features that provide the best classification accuracy of the linear tests. We applied the developed algorithm to induce the DT from the large EEG dataset consisted of 65 patients belonging to 16 age groups. In these recordings each EEG segment was represented by 72 calculated features. The DT correctly classified 80.8% of the training and 80.1% of the testing examples. Correspondingly it correctly classified 89.2% and 87.7% of the EEG recordings.***

## I. INTRODUCTION

The learning of multi-class conceptions from the electroencephalogram (EEG) data is one of important biomedical problems [1, 2, 3, 4, 5, 6, 7]. In general, the learning of the conceptions, also known as an induction of the classification models, which are able to improve the clinical interpretation of EEGs, is a complex problem. Firstly, the EEGs are strong non-stationary signals whose statistics vary in width range of values [4, 5, 6]. Secondly, the characteristics of the EEGs depend on human individuality and activity as well. To obtain under these circumstances acceptable classification accuracy, the models must be induced from the large EEG data. For this reason the learning time becomes to be crucial.

The Decision Tree (DT) methods [8, 9, 10, 11, 12, 13, 14, 15] have been successfully used for learning multi-class concepts from data. These methods assume that the input features, or the attributes, that characterize the training examples must be correlated with the goal. However some input features may be irrelevant, not correlated with the goal. As we know, using the relevant features only, an induced model is able to classify the unseen examples that have not been included in the training set. If the induced model is able to classify the new examples successfully, we can say that its generalization ability is well.

In general, methods of inducting the DTs require computational time that grows proportionally to the size of the training datasets. The computational time is also increased if the training examples are non-linearly separable.

Note also that, humans can find that the DT classification models are ease-to-understand. In particular, a DT that consists of the linear combinations of the input features is easy-to-read by experts [8, 9, 11, 12, 13].

Usually DTs consist of the nodes of two types. One is a decision node containing a test, and other is a leaf node assigned to an appropriate class. A branch of the DT represents each possible outcome of a test. An example comes at the root of the DT and follows the branches until a leaf node is reached. The name of the class at the leaf is the resulting classification.

The node can test one or more of the input features. A DT is multivariate one, if its nodes test more than one of the features. The multivariate DT is much shorter than that which tests a single feature. In addition, the reducing of the size improves DT generalization ability, i.e., the ability of the DT to classify the unseen examples.

To learn concepts presented by the numeral attributes, appropriate to the EEG data, the authors [8, 9, 12, 13, 16] suggested the multivariate DTs with Threshold Logical Units (TLU) called also perceptions. These multivariate DTs, named also oblique ones, are able to classify linear separable examples.

Using the algorithms suggested in [12, 13, 17, 18], the oblique DTs can also learn to classify the non-linear separable examples. In general, these algorithms require the computational time that grows proportionally to the numbers of the training examples, input features and classes. Nevertheless, the computational time that needs to induce the multi-class models on the large datasets becomes to be crucial, especially, if the number of the training examples is the tens of the thousands.

Note that the oblique DT algorithms exploiting the ideas of the ID3, C4.5, CART and RELIEF require computational time that grows proportionally to the squared number of the training examples [14, 15]. For large datasets, as an EEG dataset, this time is expected to be huge and for this reason we do not consider these DT induction algorithms.

In this paper we firstly introduce a new neural network based construction of the oblique DT. Secondly we introduce a DT induction method and an algorithm, which is able to select the relevant features. This algorithm performs in the acceptable time on the large datasets whose examples are not linearly separable. Thirdly we describe and analyze a real-world task of EEG classification. Finally we evaluate an ability of the neural network DT to learn a multi-class concept from the EEG data.

## II. AN INDUCTION OF OBLIQUE DECISION TREES

In this section we firstly describe an algorithm for inducting a Linear Machine DT. Then we discuss in detail a training procedure known as a Pocket Algorithm and shortly a Thermal Training Algorithm. Finally we discuss some feature selection algorithms.

*II.A. A Linear Machine Decision Tree*

A Linear Machine (LM) is a set of the $r$ linear discriminant functions that are calculated in order to assign an example to one of the $r \geq 2$ classes [16, 17]. Each internal node of the LM tests a linear combination of $m$ input variables $x_0, x_1, x_2, \ldots, x_m$, where $x_0 \equiv 1$.

Let us now introduce a $m$ feature vector $\boldsymbol{x} = (x_0, x_1, \ldots, x_m)$ and a discriminant function $g(\boldsymbol{x})$. Then the linear test at the $j$-th node has the next form:

$$g_j(\boldsymbol{x}) = \Sigma_i w_i^{\,j} x_i = \boldsymbol{w}^{j\mathrm{T}}\boldsymbol{x} > 0,\ \ i = 0, \ldots, m,\ \ j = 1, \ldots, r, \quad (1)$$

where $w_0^{\,j}, \ldots, w_m^{\,j}$ are real-valued coefficients, also known as a weight vector $\boldsymbol{w}^j$ of the $j$-th TLU.

The LM assigns an example $\boldsymbol{x}$ to the $j$ class if and only if the output of the $j$-th node is higher than the outputs of the other nodes:

$$g_j(\boldsymbol{x}) > g_k(\boldsymbol{x}),\ \ k \neq j = 1, ..., r. \quad (2)$$

Note that this strategy of making decision is known also as a Winner Take All (WTA) one.

To train a LM, the weight vectors $\boldsymbol{w}^j$ and $\boldsymbol{w}^k$ of the discriminant functions $g_j$ and $g_k$ are updated on an example $\boldsymbol{x}$ that the LM misclassifies. A learning rule increases the weights $\boldsymbol{w}^j$, where $j$ is the class to which an example $\boldsymbol{x}$ actually belongs, and decreases the weights $\boldsymbol{w}^k$, where $k$ is the class to which the LM erroneously assigns the example $\boldsymbol{x}$. This is done by using the next error correction rule

$$\boldsymbol{w}^j := \boldsymbol{w}^j + c\boldsymbol{x},\ \ \boldsymbol{w}^k := \boldsymbol{w}^k - c\boldsymbol{x}, \quad (3)$$

where $c > 0$ is amount of correction.

If the training examples are linearly separable, above procedure trains a desirable LM in a finite number of the steps [16]. If the examples are not linearly separable, a training procedure (3) cannot provide a predictable classification accuracy. For this case the other training procedures have been suggested some of them we will discuss below.

*II.B. A Pocket Algorithm*

To train the DT on the examples, which are not linearly separable, Gallant had suggested using a Pocket Algorithm [16]. This algorithm seeks the weights of a multivariate test that minimizes the number of the classification errors. The Pocket Algorithm uses an error correction rule (3) to update the weights $\boldsymbol{w}^j$ and $\boldsymbol{w}^k$ of the corresponding discriminant functions $g_j$ and $g_k$. During normal training, the algorithm saves in the pocket the best weight vectors $\boldsymbol{W}^P$ that happen.

In addition, Gallant has suggested the "ratchet" modification of the Pocket Algorithm. The idea behind this algorithm is to replace of the weight $\boldsymbol{W}^P$ by current $\boldsymbol{W}$ only if the current LM has correctly classified more training examples than that used $\boldsymbol{W}^P$. The modified algorithm finds the optimal weights if the training time was given enough.

To realize this idea, the algorithm cycles training the LM for the given number $n_e$ of the epochs. For each epoch, the algorithm counts the current length $L$ of the sequence of the examples classified correctly as well as an accuracy $A$ of the LM on all training examples.

In correspondence to an inequality (2), the LM assigns a training example $(\boldsymbol{x}, q)$ to a $j$-th class, where $q$ is a class to which an example $\boldsymbol{x}$ actually belongs. The LM training algorithm consists of the following steps:

---

1. Initialize the weights $\boldsymbol{W} := (\boldsymbol{w}^1, \ldots, \boldsymbol{w}^r)$ by random values;
2. Set the pocket parameters:
   $\boldsymbol{W}^P = \boldsymbol{W},\ \ L^P = 0,\ \ A^P = 0;$
3. **for** $i = 1$ **to** $n_e$ **do**
4. Select an example $(\boldsymbol{x}, q)$ randomly from the dataset;
5. Find a class $j$ to which the LM assigns an example $\boldsymbol{x}$;
6. **if** $j \neq q$ **then** % $\boldsymbol{x}$ misclassified
7. $L^P = 0;$
8. Update the weights $\boldsymbol{w}^j$ and $\boldsymbol{w}^k$:
   $\boldsymbol{w}^j := \boldsymbol{w}^j + c\boldsymbol{x},$
   $\boldsymbol{w}^k := \boldsymbol{w}^k - c\boldsymbol{x};$
9. **else**
10. **if** $L > L^P$ **then**
11. Evaluate the accuracy $A$;
12. **if** $A > A^P$ **then**
13. $\boldsymbol{W}^P = \boldsymbol{W},\ \ L^P = L,\ \ A^P = A;$

---

Note that the algorithm finds the best weights $\boldsymbol{W}^P$ for the time that grows proportionally to the numbers of the training examples, input variables, classes, and epochs. The number of epochs must be given enough large in order to achieve an acceptable classification accuracy of the LM. For example, in our case, the number of the epochs we have set to be maximal, equal to the number of the training examples. Note also that the best classification accuracy of the LM is achieved if the amount $c$ of correction is equal to 1.

*II.C. A Thermal Procedure*

When the training examples are not linearly separable, the classification accuracy of the LM is unpredictable. There are two cases when the behavior of the LM is destabilized during learning [17]. In the first case a misclassified example is far from a dividing hyperplane, and for removing this error the hypeplane must be substantially readjusted. Such relatively large adjustments destabilize the training procedure. In the second case a misclassified example lies very close to the dividing hyperplane, and the weights are not converged.

To improve the convergence of the training algorithm, a thermal procedure has been suggested [17]. Firstly, this procedure decreases an attention to the large errors by using the next correction

$$c = \beta/(\beta + k^2),\ \ k = (\boldsymbol{w}^j - \boldsymbol{w}^i)^{\mathrm{T}}\boldsymbol{x}/(2\boldsymbol{x}^{\mathrm{T}}\boldsymbol{x}) + \varepsilon, \quad (4)$$

where β is a calculated parameter initialized by 2, and $\varepsilon > 0.1$ is a given constant.

To handle a parameter β during training, the magnitudes of the weight vectors are summed. If the sum value decreased for the current weight adjustment, but increased during the previous adjustment, then a parameter β is reduced, $\beta := a\beta - b$, where $a$ and $b$ are the given constants.

The reducing of β enables to spend more time for training the LM with small value β that needs to refine the location of the dividing hyperplane. However, note that the experimental results of the authors [12, 14], which have implemented the thermal procedure, draw that its training time is comparable to the time of training the LM.

### *II.D. A Feature Selection*

To obtain an accurate and understandable DT, we must eliminate the features that are not able to contribute to the classification accuracy at nodes in the tree. These features may be irrelevant, corrupted by noise, or correlated with other features and often cause the DT to be over-fitted.

In this section we discuss Sequential Feature Selection (SFS) algorithms based on a greedy heuristic used to eliminate the irrelevant features. The selection is performed while the DT nodes learn that, as we know, allows to avoid the over-fitting more effectively than known methods of feature pre-processing that precede training.

A SFS algorithm exploits a bottom up search method and starts with one features then iteratively adds the new features that provide the most improvement of the quality of the linear test. The algorithm continues to add the features while a specified stopping criterion is met. During this process, the best linear test $T_b$ with the minimum number of the features is saved. In general, the SFS algorithm includes the next steps.

---

1. Run the $m$ linear tests $T_1, \ldots, T_m$ with the single feature.
2. Select the best $T_1$.
3. Set $i = 1$, $T_b = T_1$.
4. Find the best test $T_{i+1}$.
5. **if** the test $T_{i+1}$ is better than $T_b$, **then** $T_b = T_{i+1}$.
6. **if** the stopping criterion is met, **then** stop and return $T_b$.
7. **otherwise**, $i := i + 1$, and **go to** step 4.

---

The search must stop when all features have been involved in the test. In this case it is required to calculate $m + (m - 1) + \ldots + (m - i)$ linear tests, where $i$ is the number of the steps. Clearly, that if the number $m$ of the features as well as the number $n$ of the examples are large, the computational time that needs for reaching the end point may be unacceptable.

To stop the search before end point and reduce the computational time, the authors [14] suggested to use the next heuristic stopping criterion. They observed that if the accuracy of the best test is decreased by more than 10%, then the chance of finding a better test with more features is slight. Note that the classification accuracy of the resulting linear test depends on the order in which the features have been included in the test.

For the SFS algorithm, the order of including the features is determined by their classification accuracy. As we know, the accuracy depends on the initial weights as well as on the sequence of the training examples selected randomly. Subsequently, the linear test can be built non-optimally, i.e., the test can include more (a case of over-fitting) or less features than it needs to obtain the best classification accuracy. A chance of selecting the non-optimal linear test is expected to be high, because the algorithm compares two tests, which are differed by one feature only. In section 4, we will describe a new training procedure, which is able to select the relevant features at the tests more effectively.

## III. INDUCTION OF DECISION TREES FROM EEG DATA

In this section we firstly discuss the computational performance of the Pocket Algorithm used for inducting the LM from the EEG data. Then we describe a new neural network based structure of the LM and discuss some advantages of this scheme.

### *III.A. A Performance of Pocket Algorithm*

The EEG data that we need to classify are split on the several classes that are extremely overlapped each other. The centers of these classes are very close to each other. This circumstance worsens LM performance dramatically. We found experimentally that the LM failed learning the EEG data. Below we shortly discuss some possible reasons.

The Pocket and Thermal Algorithms, as we know, update the weights of two linear tests on a misclassified example. The updating of the weights then initiates setting the size $L^p$ of the example sequence to 0, at least, we should do it in our implementation of the algorithm.

What happens on the next examples? If the next example is correctly classified and $L > L^P$, the algorithm will evaluate the accuracy $A$ of the LM on all training examples. If the next example will be also correctly classified, the algorithm again will evaluate the accuracy of the LM on the training set. We can see that calculation time grows quickly at these beginning steps, especially if the number of training examples is large.

The misclassification of the example causes updating the LM weights and setting $L^P = 0$. Therefore for the next example $L > L^P$ and the algorithm will again evaluate the accuracy of the updated LM on all training examples. This will be happens often because the training examples are heavily overlapped and not linear separable.

We can conclude that for these reasons above training algorithm did not able to induce the LM from the large EEG data for an acceptable time. If the training data consist of the examples that are heavily overlapped, the algorithm requires extensive computational time.

*III.B. A Neural Network Decision Tree*

The idea behind our algorithm for inducting DTs is to individually train their TLUs and group them in order to linearly approximate the desirable dividing surfaces. A TLU that realizes a DT linear test is individually trained to classify the examples of two classes. For $r$ classes therefore it is needed to compose the $\binom{r}{2}$ variants of the training subsets and train the same number of the TLUs, where $\binom{r}{2} = r(r-1)/2$. The trained TLUs that deal with one class are taken out at one group, or a neural network, so that the number of the groups corresponds to the number $r$ of the classes. The TLUs that belong to one group are superposed in order to linearly approximate a dividing surfaces of the corresponding classes.

Let $f_{i/j}$ be a TLU that performs a linear test. The TLU $f_{i/j}$ learns to divide the examples belonging to the $i$-th and $j$-th classes $\Omega_i$ and $\Omega_j$. If the training examples are linearly separable, then the output $y$ of the TLU is

$$y = f_{i/j}(\boldsymbol{x}) = 1, \forall\, \boldsymbol{x} \in \Omega_i, \quad y = f_{i/j}(\boldsymbol{x}) = -1, \forall\, \boldsymbol{x} \in \Omega_j, \qquad (5)$$

To explain an idea of our algorithm, in Fig. 1 we depicted $r$ = 3 overlapping classes $\Omega_1$, $\Omega_2$ and $\Omega_3$, whose centers are $C_1$, $C_2$ and $C_3$. The number of the TLUs therefore is equal to $\binom{r}{2}$ = 3. In Fig. 1 the three lines $f_{1/2}$, $f_{1/3}$ and $f_{2/3}$ depict the hyperplanes of the TLUs trained to divide the examples of two classes: $\Omega_1$ and $\Omega_2$, $\Omega_1$ and $\Omega_3$, $\Omega_2$ and $\Omega_3$, respectively.

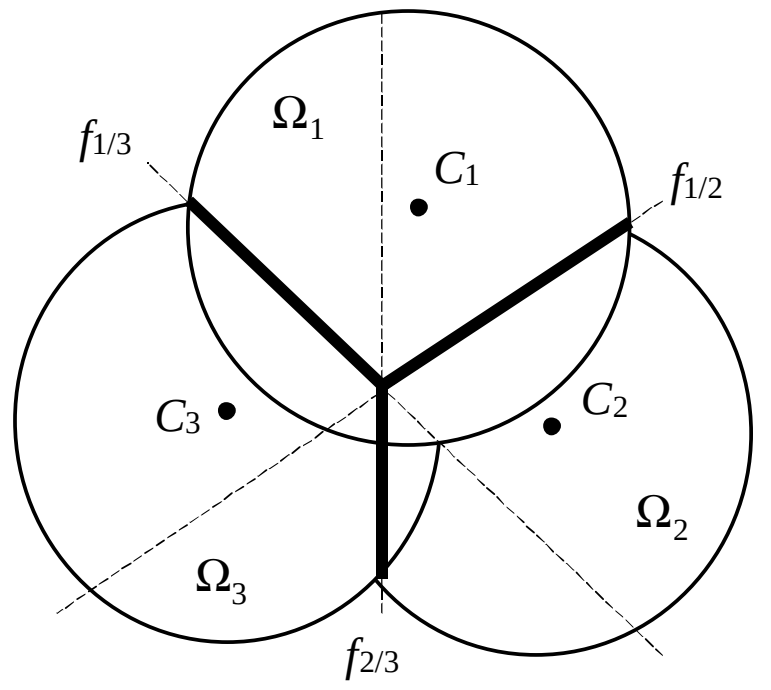

Fig. 1: Hypeplanes $f_{1/2}$, $f_{1/3}$ and $f_{2/3}$ dividing classes $\Omega_1$, $\Omega_2$ and $\Omega_3$

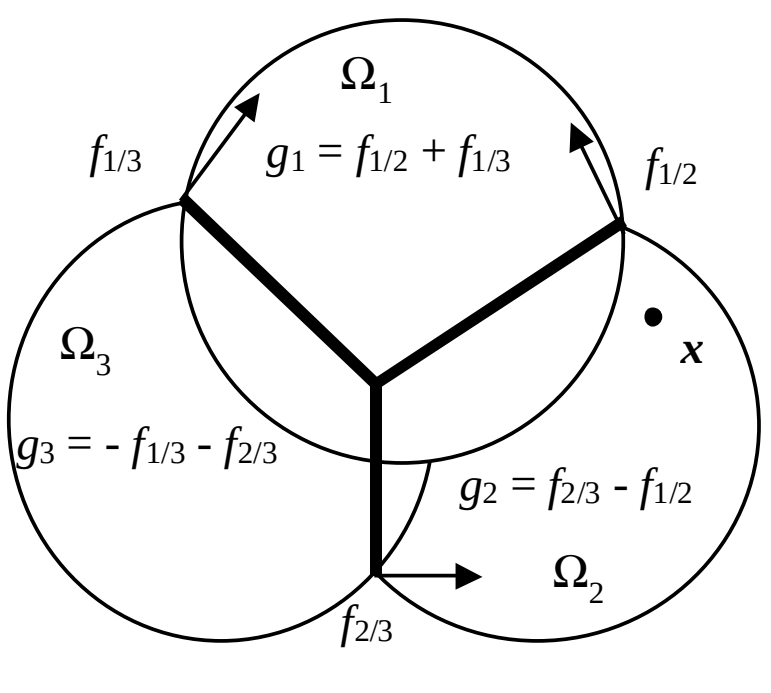

Fig. 2: The approximation by the dividing surfaces $g_1$, $g_2$ and $g_3$

In Fig 2 we depicted three new dividing surfaces $g_1$, $g_2$ and $g_3$ of the corresponding classes. The first surface $g_1$ is a superposition of the linear tests $f_{1/2}$ and $f_{1/3}$, i.e., $g_1 = f_{1/2} + f_{1/3}$. These tests are summarized with weights equaled to 1 because both tests $f_{1/2}$ and $f_{1/3}$ give the positive outputs on the examples of the class $\Omega_1$. Correspondingly, the second and third surfaces are $g_2 = f_{2/3} - f_{1/2}$ and $g_3 = -f_{1/3} - f_{2/3}$.

We can see that an example $\boldsymbol{x}$ that belongs to a class $\Omega_2$ causes the outputs of $g_1$, $g_2$ and $g_3$ to be equal to 0, 2 and –2 correspondingly:

$$\begin{aligned} g_1(\boldsymbol{x}) &= f_{1/2}(\boldsymbol{x}) + f_{1/3}(\boldsymbol{x}) = 1 - 1 = 0, \\ g_2(\boldsymbol{x}) &= f_{2/3}(\boldsymbol{x}) - f_{1/2}(\boldsymbol{x}) = 1 + 1 = 2, \\ g_3(\boldsymbol{x}) &= -f_{1/3}(\boldsymbol{x}) - f_{2/3}(\boldsymbol{x}) = -1 - 1 = -2. \end{aligned} \qquad (6)$$

We can see that the WTA strategy that the DT uses does correctly assign an example $\boldsymbol{x}$ to the class $\Omega_2$.

To approximate the dividing surfaces $g_1, \ldots, g_r$, we suggest to exploit two-layer feed-forward neural networks consisting of the TLUs. It is easy to see that $g_1$, $g_2$ and $g_3$ can be approximated by a neural network that consists of the input, hidden and output layers, see Fig. 3.

In Fig. 3, we depicted three hidden neurons that perform the linear tests $f_{1/2}$, $f_{1/3}$ and $f_{2/3}$, whose weights are $\boldsymbol{w}^1$, $\boldsymbol{w}^2$ and $\boldsymbol{w}^3$. The hidden neurons connected to the output neurons $g_1$, $g_2$ and $g_3$ with the weights equal to (+1, +1), (–1, +1) and (–1, –1), respectively.

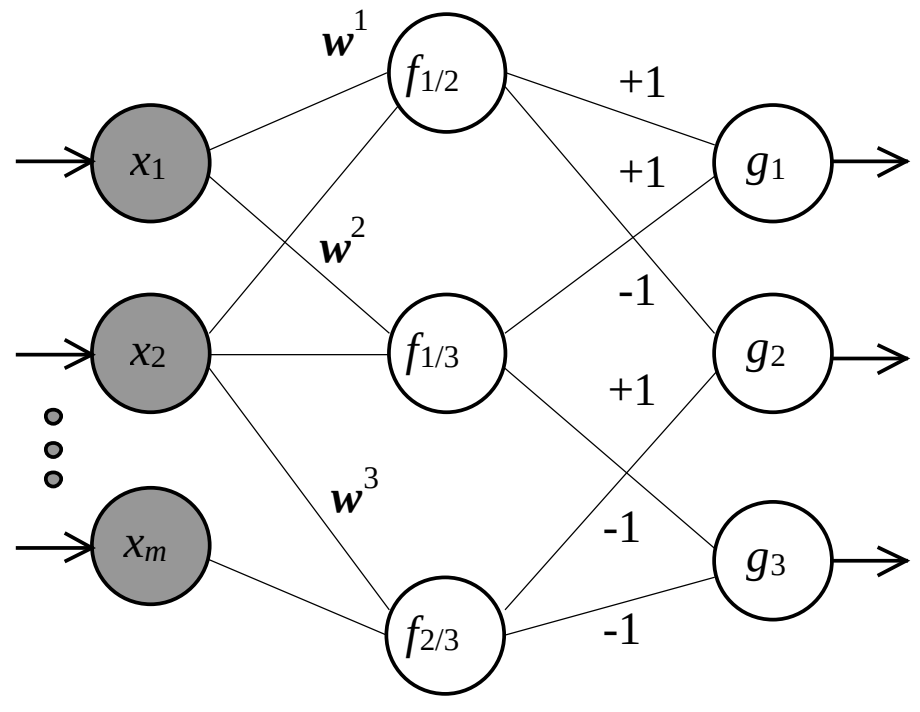

Fig 3: An example of the neural network decision tree

In general case for $r > 2$ classes, the neural network consists of $r(r-1)/2$ hidden neurons $f_{1/2}, \ldots, f_{i/j}, \ldots, f_{r-1/r}$ and $r$ output neurons $g_1, \ldots, g_r$, where $i < j = 2, \ldots, r$. The output neuron $g_i$ connected to the $(r-1)$ hidden neurons that are partitioned by two groups. The first group consists of the hidden neurons $f_{i/k}$ for which $k > i$. The other group consists of the hidden neurons $f_{k/i}$ for which $k < i$. Since the outputs of the hidden neurons are described by an equation (5), the weights of the output neuron $g_i$ connected to the hidden neurons $f_{i/k}$ and $f_{k/i}$ are equal to + 1 and –1 correspondingly.

The connectivity of the hidden neurons performing the linear tests depends on the significance of the features $x_1, \ldots, x_m$. In the next section we will discuss the training algorithm that is able to select the relevant features.

## IV. A TRAINING ALGORITHM OF LINEAR TESTS

In this section we firstly describe an algorithm for training the weights of the linear tests. Secondly we describe in detail an algorithm that is able to select the relevant features during training the linear tests.

### *IV.A. Training of Weights*

Within the neural network DT introduced above, the hidden neurons perform the linear tests. Methods that can be used for training the linear tests depend on distortions and noise in the training data.

In our case of real EEG data, the distortion and noise are approximately distributed by a skew Gauss distribution function. We experimentally found that the skewness of the distribution function is weak. For this reason, we suggest to use a standard Least Square Error (LSE) technique. This technique is typically used to fit linear discriminant functions when the training examples are linear separable. In general, LSE technique has been effectively used for training the weights of the multivariate functions [19, 20] as well as the TLUs.

To shortly describe a LSE training procedure let us introduce a $(n\times m)$ training matrix $\mathbf{X}$ and a target $(n\times 1)$ vector $\boldsymbol{t}$, where $n$ is the number of the training examples, and $m$ is the number of the variables, a linear multivariate test uses. Note that $m \leq m_0$, where $m_0$ is the number of all input variables. The number of the examples must exceed the number of the variables (the number of the unknown weights), i.e., $n > m$.

A matrix $\mathbf{X}$ contains the training examples that belong to the $i$-th and $j$-th classes, $i \neq j = 1, \ldots, r$. The elements of the target vector $\boldsymbol{t}$ are marked by 1 and –1 for the examples belonging to the $i$ and $j$ classes respectively.

The output $y$ of the linear test is described by (5). A linear test assigns an example $\boldsymbol{x}$ to the $i$-th class if $y > 0$, otherwise an example is assigned to the $j$-th class. A minimal training error is achieved, if the LSE procedure fits the weight vector $\boldsymbol{w}$ of the linear test so that:

$$e = \| \mathbf{X}\boldsymbol{w} - \boldsymbol{t} \| \rightarrow min, \qquad (7)$$

where $e$ is a residual squared error of the linear test on the training set $\mathbf{X}$, and $\| \cdot \|$ is the Euclidian norm.

Thus for linear separable data the LSE procedure may yield the desirable weight vector $\boldsymbol{w}^*$ that minimizes the residual error $e$ (7). In this case the classification accuracy of the linear test using weights $\boldsymbol{w}^*$ depends on the level of the distortions and noise in the training data.

### *IV.B. A Selection of Relevant Features*

A feature selection algorithm based on a bottom up search, we discussed in section 2. This algorithm searches for the new features that cause the largest increasing the classification accuracy of the linear test. We can see that the tests the algorithm compares are differed in one feature. In addition, the greedy heuristic exploited always adds a feature that provides the largest increasing the accuracy of the test.

We experimentally found firstly that a chance of building the best linear test is significantly increased if the comparable tests are differed in more than one features. Secondly, in real EEG data distorted by noise the greedy heuristic often catches a local maximum in the classification accuracy. In addition to these heuristics, our training algorithm uses the examples that not been used for fitting the weights of the linear tests. The use of these examples allows to improve the selection of the relevant features and avoid the over-fitting the linear tests more effectively.

In general, appropriate training algorithm we developed is able to select the relevant features includes the next steps.

---

1. Run $m$ linear tests with the single features;
2. Create the pools $P$ and $F$;
3. Set a best test $T_b = \varnothing$;
4. Set $A_b = 0$;
5. **for** $k = 1$ to $N_t$ **do**
6. Set a test $T = \varnothing$;
7. Set $nof = 0$; % the number of the features
8. Set $A = 0$;
9. Set $i = 1$; % the feature index in $F$
10. **if** the feature $x_i$ is added **then**
11. $T_1 := [T \ x_i]$; % a candidate-test
12. Train the weights and calculate the accuracy $A_1$ of $T_1$;
13. **if** $A_1 > A$ **then**
14. $T = T_1$, $A = A_1$;
15. $nof := nof + 1$;
16. **if** $A > A_b$ **then**
17. $T_b = T$; $A_b = A$;
18. **if** the stopping criterion is met **then**
19. Return a best test $T_b$;
20. **else** $i := i + 1$, **go to** step 10;

---

To realize a bottom up search, we exploit some techniques that are typical for evolutionary algorithms. The algorithm starts to train the liner tests which contain one feature $x_i$, $i = 1, \ldots, m$. Then for each of $m$ unit-variate tests, the algorithm calculates the probability $p$

$$p_i = A_i/\Sigma_j^m A_j, \quad i = 1, \ldots, m, \qquad (8)$$

where $A_i$ is the classification accuracy of the $i$-th test.

The calculated values of the probabilities are arranged in the descent order, i.e., $p_{i1} \geq p_{i2} \geq \ldots \geq p_{im}$, and then saved in a pool $P$. In accordance with arranging indexes $i_1, i_2, \ldots i_m$ the features $x_{i1}, x_{i2}, \ldots, x_{im}$ are saved in a pool $F$.

At the next steps, the algorithm sets the test $T := \varnothing$ and the index $i = 1$, and then it attempts to add the feature $x_i$ to $T$. If this is happened with a probability $p_i$, a candidate-test $T_1$ is formed and its weights are fitted by the LSE procedure that minimizes the residual error (7). Then the classification accuracy $A_1$ of the test $T_1$ is calculated.

If the accuracy $A_1$ becomes to be higher than the accuracy $A$ of the current test $T$, then a candidate-test $T_1$ replaces a test $T$. The number of the features the new liner test $T_1$ includes is increased by 1.

The algorithm is repeated until a stopping criterion is met. The stopping criterion is met in two cases: firstly, when the linear test includes the given number $N_f$ of the features, secondly, when all features have been tested.

Note that at the 16 step, the algorithm compares the linear tests $A$ and $A_b$ that have different features. This increases a chance to search out the best linear test considerably.

As a result, a unique set of the features is formed. Using these features, the TLU classifies the examples well. To enlarge a chance of finding a best solution, the tests are multiply trained by the given number $N_t$ attempts, each time with the different sequence of the features.

Note that for fitting the weights of the TLUs we used 2/3 of all training examples. The classification accuracy of the TLUs was evaluated on all training examples. The number $N_t$ of the attempts we varied from 5 to 25. For 72 features, the maximal number $N_f$ of the features that the linear test can include was equal to 60.

## V. A STRUCTURE AND ANALYSIS OF EEG DATA

In this section we describe the structure of real EEG data we used to train a neural network DT. Then we discuss some statistical characteristics of the EEG data.

### *V.A. A Structure of EEG Data*

We used the EEG data that was recorded via the standard channels C3 and C4 from 65 patients belonging to different age groups. Following [4, 5, 6], each EEG segment in these recordings has been represented by 72 calculated features. The first of them are based on the features that are the power spectral densities calculated on 10-second interval into 6 frequency bands: sub-delta (0-1.5 Hz), delta (1.5-3.5 Hz), theta (3.5-7.5 Hz), alpha (7.5-13.5 Hz), beta 1 (13.5-19.5 Hz), and beta 2 (19.5-20 Hz). These densities were calculated for the channels C3 and C4 as well as for their total sum. The second features are the relative and absolute power densities and the variances of the first variables. The EEG data were finally normalized.

The EEG-viewer has manually deleted the artifacts from these recordings and then assigned the normal segments to the 16 classes in accordance with the ages of the patients. Note that after cleaning an average rate of the outlying segments did not exceed 6%, and the total sum of all EEG segments was equal to 59,069.

The aim of our experiment was to induce a multi-class model from the large EEG dataset. We would like to use this model in order to determine the EEG maturation of the patients. The index of the maturation is used for clinical diagnosis of some pathologies [5, 6, 7]. Since a desirable model must be not patient-depended, we should induce the DT from the large EEG dataset.

### *V.B. A Statistical Analysis of EEG Data*

Firstly, we remind that the EEG is heavily non-stationary signal, because its spectral parameters vary over time. To demonstrate this, we calculated two principal components $p_1$ and $p_2$ on the EEG recordings made for two patients belonging to different classes.

In Fig 4 we depicted in a space of two principle components $p_1$ and $p_2$ the EEG segments belonging to four time intervals: 1:300, 301:600, 601:900, and 901:1200.

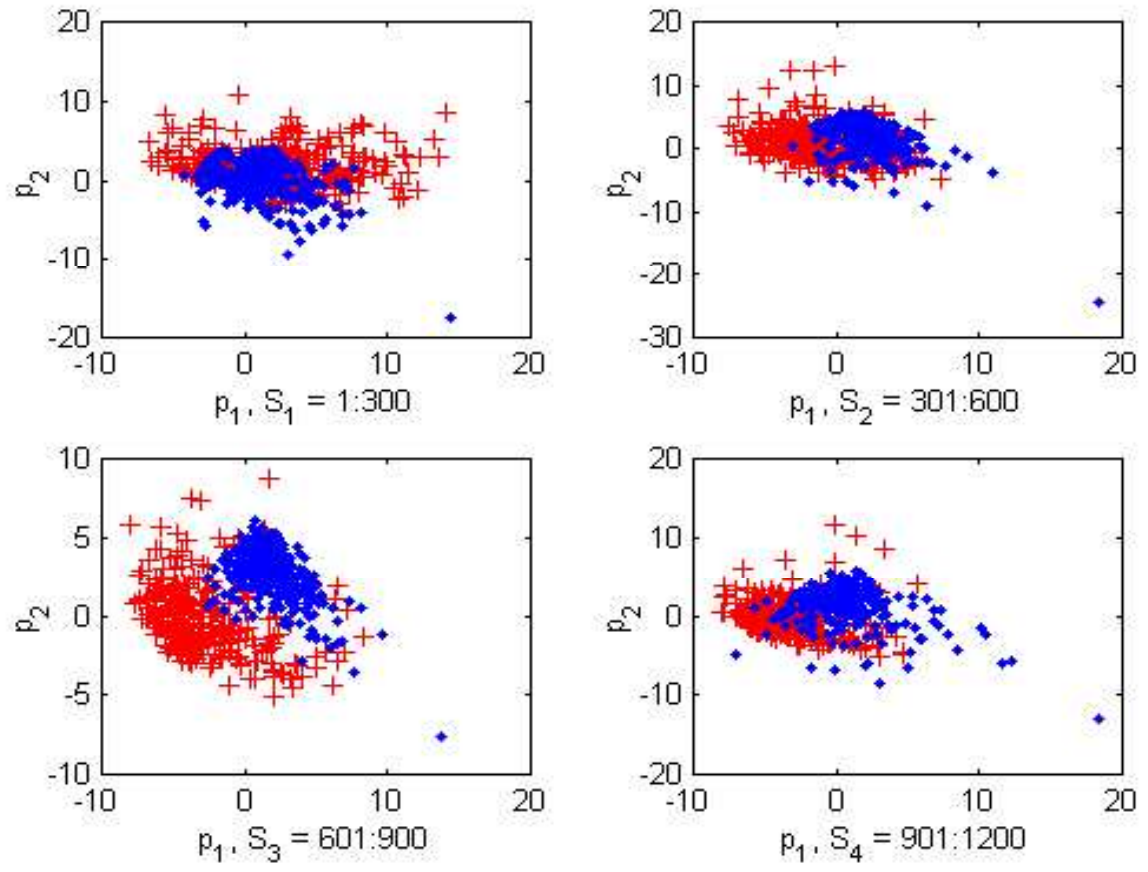

Fig 4 The EEG segments of two patients into four intervals

We can see that the values $p_1$ and $p_2$ calculated into these intervals vary over time considerably. Such a variability does not allow distinguishing the EEG segments belonging to different classes by using these principle components.

Secondly, we must remind that the EEGs reflect individual activity of patients. This activity has a chaotic character that significantly increases the group variance of the class. To analyze the effect of the individual activity, we used the next statistical technique.

Let us introduce a variance $v(x_j)$ of all $r$ classes and a group variance $s_i(x_j)$ of the $i$-th class:

$$v(x_j) = \frac{1}{r}\sum_{k=1}^{r}(\tilde{x}_{kj} - \tilde{y}_j)^2,\quad \tilde{y}_j = \frac{1}{r}\sum_{k=1}^{r}\tilde{x}_{kj}, \tag{9}$$

$$s_i(x_j) = \frac{1}{N}\sum_{k=1}^{N}(x_{kj} - \tilde{x}_{ij})^2,\quad \tilde{x}_{ij} = \frac{1}{N}\sum_{k=1}^{N}x_{kj}, \tag{10}$$

where $N$ is the number of the segments into the $i$-th class, $x_{kj}$ is a $j$-th feature value of a $k$-th example, $\tilde{x}_{ij}$ is a mean value of a $j$-th feature calculated for the $i$-th class.

Then we can evaluate a significance of a $j$-th feature as the next ratio:

$$d_j = \alpha \frac{v(x_j)}{\sum_{i=1}^{r} s_i(x_j)},\quad j = 1, \ldots, m, \tag{11}$$

where $\alpha = 100$ is a coefficient.

We see that the value $d_j$ is decreased proportionally to the sum of the group variances $s_i$ and increased proportionally to the variance $v(x_j)$ of the classes. Clear that the group variance $s_i$ grows proportionally to an individual activity of the patients that belong to the same group. Therefore we can conclude that $d_j$ is a measure of the significance of the feature $x_j$.

Fig. 5 depicts the values of *d*, *v* and *s* for all 72 features calculated on the training set. We can see that the most relevant feature is $x_{36}$, and less relevant one is $x_{38}$.

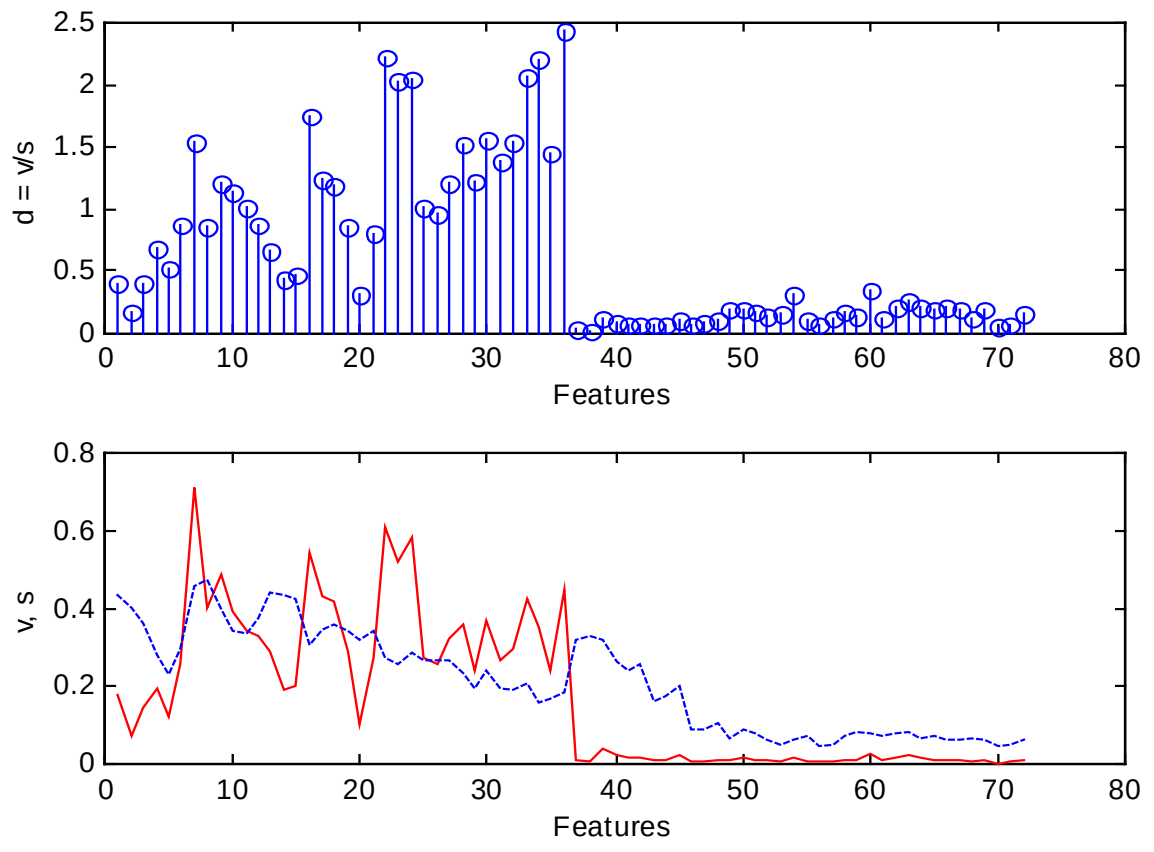

Fig 5: The significance of all 72 features

In Fig. 6 we depicted the intervals 3sigma calculated for the feature $x_{36}$ and $x_{38}$. Observing their behavior on 16 classes, we can see that a less relevant feature $x_{38}$ is slightly depended on the classes.

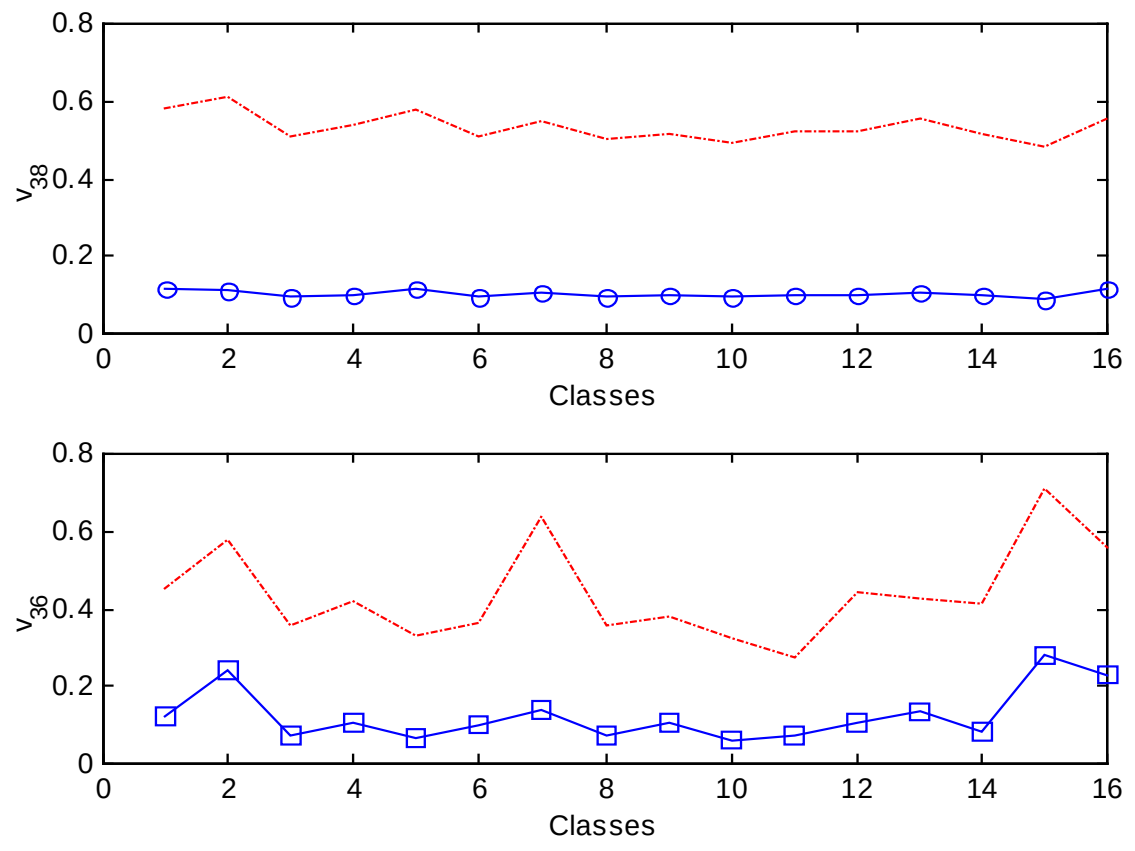

Fig 6: The intervals 3·sigma for the features $x_{38}$ and $x_{36}$

In addition, the interval 3sigma of a feature $x_{38}$ is larger than the corresponding interval of a feature $x_{36}$. However the feature $x_{36}$, as we can see, does not allow to distinguish all 16 classes properly because its interval 3sigma is yet large. On the other words, the values $x_{36}$ is expected to be almost the same on all 16 classes.

## VI. EXPERIMENTS AND RESULTS

For learning a multi-class concept from the EEG data, we used a neural network DT described in sections 3 and 4. For training we used 2/3 of all examples or the 39,399 EEG segments. The rest 19,670 segments we used for testing a trained DT. For $r$ = 16 classes, the neural network DT includes the $r$ $(r - 1)/2$ = 120 linear tests or the TLU classifiers. The training errors of these classifiers vary from 0 to 15% depicted in Fig. 7a. Note that each trained classifier includes the different features. The number of the features the classifiers use varies from 7 to 58 that depicted in Fig 7b.

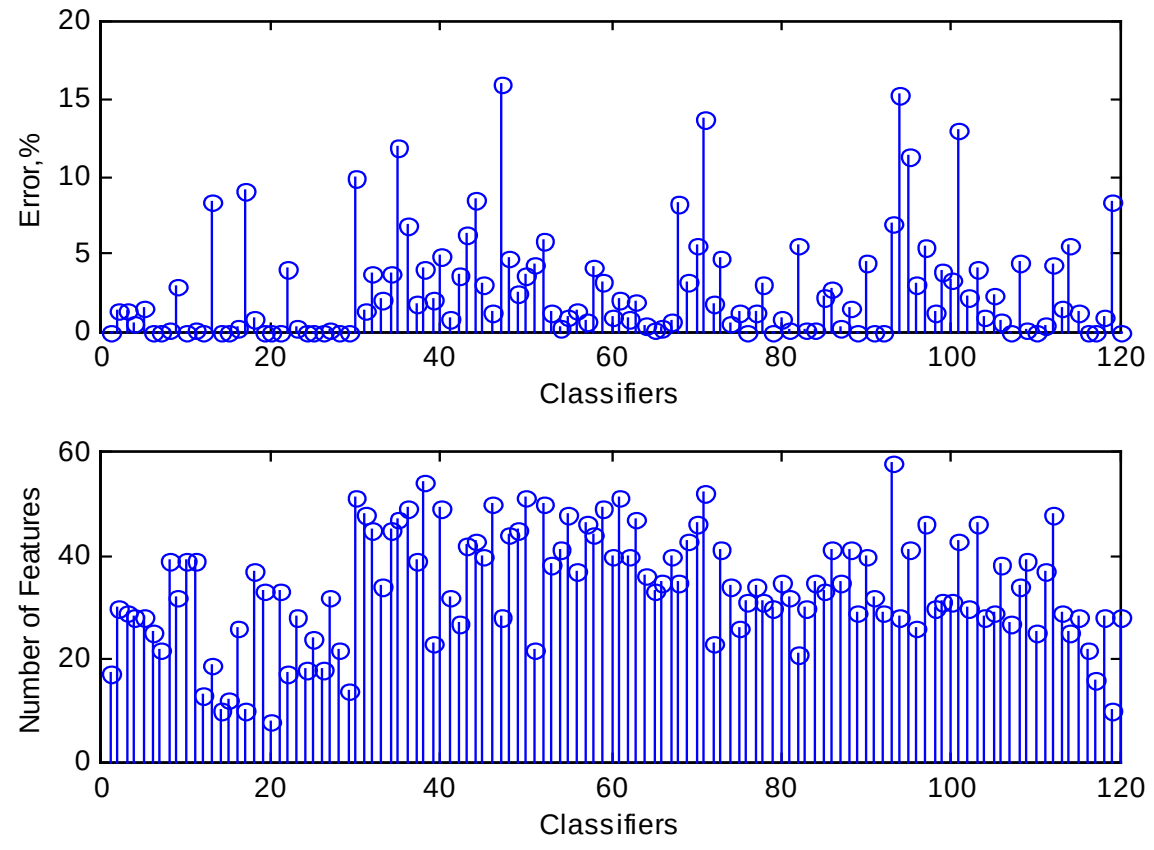

Fig 7: The training errors (a) and the number of features (b) for 120 TLU classifiers

The trained neural network DT correctly classified the 80.8% of the training and 80.1% of the testing examples that represent 65 EEG recordings. Summing all segments that belong to one EEG recording, we can conclude that the trained DT correctly classified 89.2% and 87.7% of 65 EEG recordings on the training and testing examples respectively. In Fig. 8a and 8b we depicted the summed outputs of the trained DT calculated on the testing segments of two patients respectively.

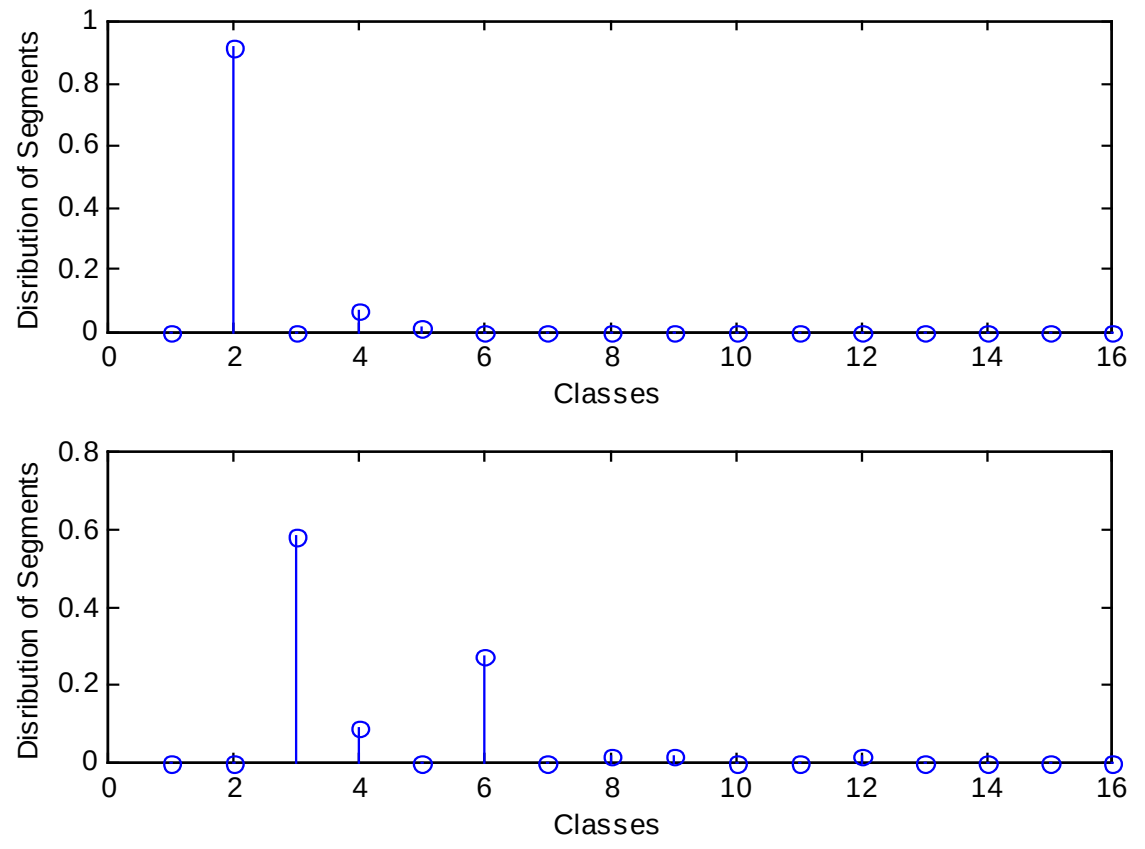

Fig 8: The summed outputs of the DT for two patients

Clearly, that the summed outputs may interpret as a distribution of the segments on all 16 classes. In this case, we can provide a probabilistic interpretation of making decisions: for example, we assigned two patients to 2 and 3 classes with probabilities 91.7 and 58 respectively, see Fig. 8.

To compare our DT induction algorithm, we applied some standard neural networks and data mining techniques to the EEG data. Firstly, we tried to train a feed-forward neural network consisting of the hidden and output layers by a fast back-propagation method by Levenberg-Marquardt the MATLAB provides. The input and output layers included 72 units and 16 neurons respectively. Secondly we trained independently $r$ = 16 binary classifiers to distinguish one class from the others. Thirdly we trained a binary decision tree that consisted of $r$ – 1 = 15 linear classifies. However all of them as well as the LM discussed in section 2 failed induction of the classification models.

## VII. CONCLUSION

For learning of multi-class conceptions from the EEG data that may be represented by the irrelevant features, we developed a neural network DT and a new training algorithm. We found that the known approaches failed induction of the classification models from the EEG data whose examples are strongly overlapped.

A neural network DT we developed consists of TLUs that perform the linear multivariate tests. A new algorithm of inducting the DTs trains the TLUs and then groups them in order to linearly approximate a desirable dividing surfaces. Each TLU is individually trained to classify the examples of two classes. The trained TLUs that deal with one class are collected at a neural network, so that the number of the groups corresponds to the number $r$ of the classes. The TLUs of one group are superposed in order to linearly approximate a dividing surface of the corresponding classes.

The training algorithm exploits a bottom up search to select the features that provides the best classification accuracy of the linear tests. For evaluating the feature significance, the algorithm uses the unseen examples, which have not been used for training the weights of the linear tests.

We applied the developed algorithm to induce the DT from the large EEG dataset that consists of 65 recordings belonging to 16 different groups. In these recordings each EEG segment was represented by 72 calculated features. For training we used 2/3 of all examples or 39,399 EEG segments. The rest of 19,670 segments we used for testing the trained DT. Finally the trained neural network DT correctly classified the 80.8% of the training and 80.1% of the testing examples. It corresponds to the correct classification of 89.2% and 87.7% from the 65 EEG recordings.

We can conclude that the new method developed for inducting neural network DTs is performed on the large EEG data successfully. We hope the method maybe used to induce classification models from large data that can be represented by irrelevant features and overlapping classes.

The authors are grateful to Frank Pasemann and Joachim Schult from TheoriLabor for enlightening discussions, Joachim Frenzel and Burghart Scheidt, Pediatric Clinic of the University Jena, for making available their EEG recordings.